\documentclass{article}
\usepackage{spconf,amsmath,graphicx}

\usepackage{enumitem}
\setlist{nosep, leftmargin=14pt}
\usepackage{hyperref}

\usepackage{mwe} 

\newcommand\myeq{\mathrel{\stackrel{\makebox[0pt]{\mbox{\normalfont\tiny def}}}{=}}}
\title{Top-k maximum intensity projection priors for\\ 3D liver vessel segmentation}
\name{Xiaotong Zhang, Alexander Broersen, Gonnie CM van Erp, Silvia L. Pintea, Jouke Dijkstra}
\address{Radiology Department, Leiden University Medical Center, Leiden, the Netherlands} 

\usepackage{comment}
\usepackage{todonotes}
\usepackage{blindtext}
\usepackage{times}
\usepackage{epsfig}
\usepackage{multirow}
\usepackage{multicol}
\usepackage{colortbl}
\usepackage{booktabs}

\newcommand{\Eq}[1]{Eq.~(\ref{eq:#1})}

\newcommand{\fig}[1]{Fig.~\ref{fig:#1}}

\usepackage{xspace}
\newcommand{\latinphrase}[1]{\textit{#1}}  

\newcommand{\ie}{\latinphrase{i.e.}\xspace}

\begin{document}
%
\maketitle

\maketitle

\begin{abstract}
Liver-vessel segmentation is an essential task in the pre-operative planning of liver resection.
State-of-the-art 2$D$ or 3$D$ convolution-based methods focusing on liver vessel segmentation on 2$D$ CT cross-sectional views, which do not take into account the global liver-vessel topology.  
To maintain this global vessel topology, we rely on the underlying physics used in the CT reconstruction process, and apply this to liver-vessel segmentation.
Concretely, we introduce the concept of \textsl{top-k maximum intensity projections}, which mimics the CT reconstruction by replacing the integral along each projection direction, with keeping the top-k maxima along each projection direction. 
We use these top-k maximum projections to condition a diffusion model and generate 3$D$ liver-vessel trees.
We evaluate our 3$D$ liver-vessel segmentation on the \textsl{3D-ircadb-01} dataset, and achieve the highest \textsl{Dice} coefficient, intersection-over-union (\textsl{IoU}), and \textsl{Sensitivity} scores compared to prior work.
\end{abstract}
\begin{keywords}
Maximum intensity projection, 3$D$ liver-vessel segmentation, CT segmentation.
\end{keywords}
\section{Introduction}
\label{sec:intro}
Liver resection is the standard surgical treatment option for primary and secondary liver cancer \cite{b1}. 
In the preoperative planning of the tumor resection, it is essential to minimize perioperative blood loss \cite{b1}.  
Due to the complex anatomy of liver vessels, it is challenging to obtain accurate liver-vessel segmentations automatically.
\begin{figure}[t!]
  \centering
  \begin{tabular}{c}
    \includegraphics[width=0.75\linewidth]{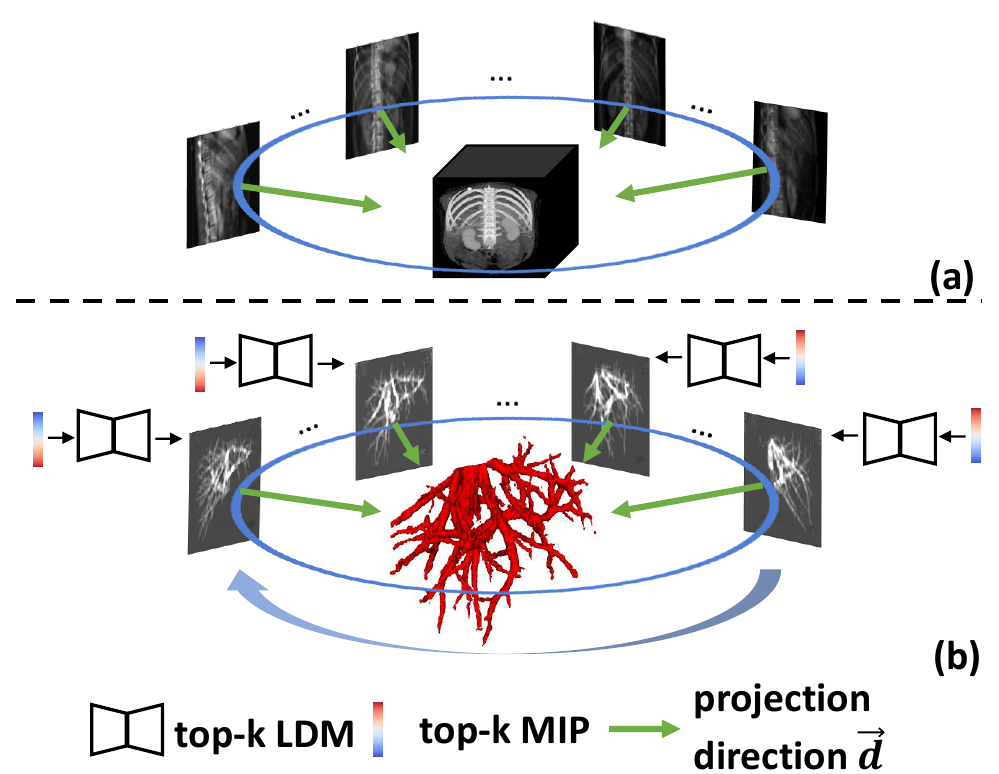} %
  \end{tabular}
  \caption{\small
    \textbf{(a) Standard CT reconstruction:} Given the integral projections and projections directions $\overrightarrow{d}$, it reconstructs the underlying  3$D$ object by back-projecting the integral along each direction;
    \textbf{(b) Our proposed \textsl{top-k MIP}:} Given the projections directions $\overrightarrow{d}$, our model reconstructs the 3$D$ liver-vessel tree by computing the top-k maximum value of the CT scan along each direction and inputting this into a latent diffusion model. 
  }
  \label{fig:summary}
  \vspace{-10px}
\end{figure}

Liver-vessel segmentation is currently done with convolutional neural networks (CNN). 
Prior work relies on 2$D$ CNN \cite{b4,b5} or 3$D$ CNN \cite{b6,b7} methods to perform segmentation on CT images. 
However, these methods use sub-volumes (cropped in three dimensions) of the CT images. 
These sub-volumes approaches do not take the global topology of the liver-vessel tree into account which could lead to discontinuous and incomplete vessel tree predictions. 

To address these shortcomings, prior work \cite{b11,b14,b15} makes use of \textsl{maximum intensity projection} (MIP) \cite{b10}.
Given a 3$D$ volume and a set of directions, \textsl{MIP} computes the maximum voxel value along each direction. 
Therefore, \textsl{MIP} encodes the global vessel topology on 2$D$ projections, and is characterized by high signal-to-noise ratio, while enhancing local vessel probability \cite{b10}.
However, due to the lack of 3$D$ information, \textsl{MIP} is only an adjunct for segmentation in existing \textsl{MIP}-based methods, which must be combined with a 3$D$ U-Net model \cite{b11,b14,b15}. 

\begin{figure*}[t!]
  \centering
  \includegraphics[width=.9\textwidth]{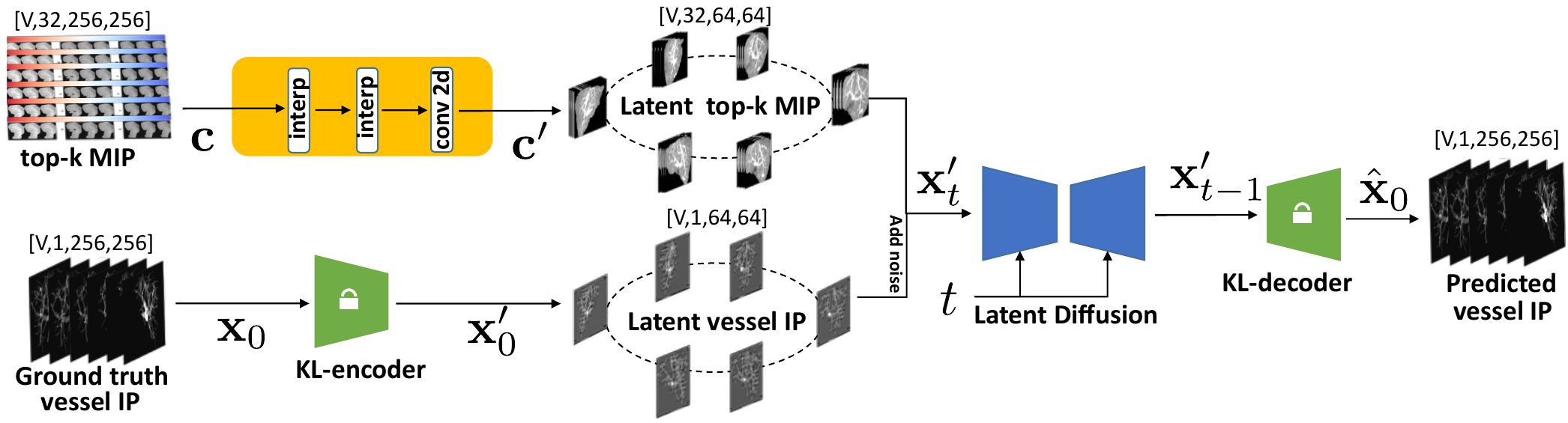}  
  \caption{\small
  \textbf{Model outline.}
  Our model represents the global 3$D$ topology of the liver by computing the \textsl{top-k MIP} over the CT volume, $\mathbf{c}$.
  Subsequently, it encodes these \textsl{top-k MIP} into a latent condition, $\mathbf{c}^\prime$ (orange). 
  This $\mathbf{c}^\prime$ latent is used to condition the latent diffusion model (blue) which recovers the ground truth vessel tree ${\mathbf{x}}_0$ from noisy inputs ${\mathbf{x}}_t$.
  We represent the ground truth via \textsl{integral projections (IP)} of the 3$D$ ground truth vessel tree.
  The ground truth ${\mathbf{x}}_0$ and the noisy input ${\mathbf{x}}_t$ are encoded via a KL-autoencoder (green) to be used in the latent diffusion U-net \cite{b19}.
  We denote the different viewing directions by $V$ in the batch size.
  }
  \label{fig:network}
  \vspace{-5px}
\end{figure*}
To incorporate the 3$D$ liver-vessel structure, we introduce a \textsl{top-k maximum intensity projection (\textsl{top-k MIP})}. 
Unlike \textsl{MIP}, which preserves only the maximum voxel value along any given direction, a \textsl{top-k MIP} preserves the top-\textsl{k} maximum values along that direction.
This allows us to encode fine-grained information regarding the underlying 3$D$ structure of the vessel tree. 
This information would be missed by simply taking the maximum. 
Therefore, our model no longer needs to rely on additional 3$D$ convolutions over the CT cross-sectional view to encode this 3$D$ structure.

To sum up, we focus on 3$D$ liver-vessel segmentation by mimicking the process of CT reconstruction \cite{b17}, as shown in \fig{summary}(a).
In \fig{summary}(b) we propose the \textsl{top-k MIP} based 3$D$ vessel segmentation.
To reconstruct the 3$D$ liver-vessel tree use these \textsl{top-k MIP}s as conditions in a latent diffusion model (LDM) \cite{b19}.
Overall, we make the following contributions:
(1) we propose incorporating the 3$D$ topology of the liver structure in a principled way by relying on the underlying physics used in CT reconstruction;
(2) to this end, we propose a novel \textsl{top-k maximum intensity projection} that encodes the fine-grained 3$D$ liver-vessel structure and combine this into a latent diffusion model;
(3) finally, we demonstrate improved accuracy of our model on the \textsl{3D-ircadb-01} \cite{b23} dataset when compared to state-of-the-art methods such \textsl{nnUNet} \cite{b24}, \textsl{SwinUNetr} \cite{b25}, \textsl{EnsemDiff} \cite{b26} and \textsl{MedSegDiff} \cite{b27}.



\section{Top-k maximum intensity projection for liver-vessel segmentation}
\label{sec:method}
Our model, displayed in \fig{network}, represents the global 3$D$ topological structure of the liver via the \textsl{top-k MIP} (denoted by $\mathbf{c}$) of the CT cross-sectional view. 
We further encode \textsl{top-k MIP} (via the yellow block) into the latent condition, $\mathbf{c}^\prime$.
This $\mathbf{c}^\prime$ is used to condition the latent diffusion model (depicted in blue) which generates the integral projections of ground truth vessel tree by denoising a given noisy ground truth, in the latent space.
The ground truth, ${\mathbf{x}}_0$, and the noisy input $\mathbf{x}_t$ are represented as \textsl{integral projections} over directions $\overrightarrow{d}$ of the true 3$D$ vessel tree segmentation.
These \textsl{integral projections (IP)} are further encoded\slash decoded via an autoencoder (depicted in green), to be used in the latent diffusion model.

\subsection{Ground truth \textsl{IP} and \textsl{top-k MIP} condition}
\begin{figure}[t]
  \centering
  \includegraphics[width=0.95\linewidth]{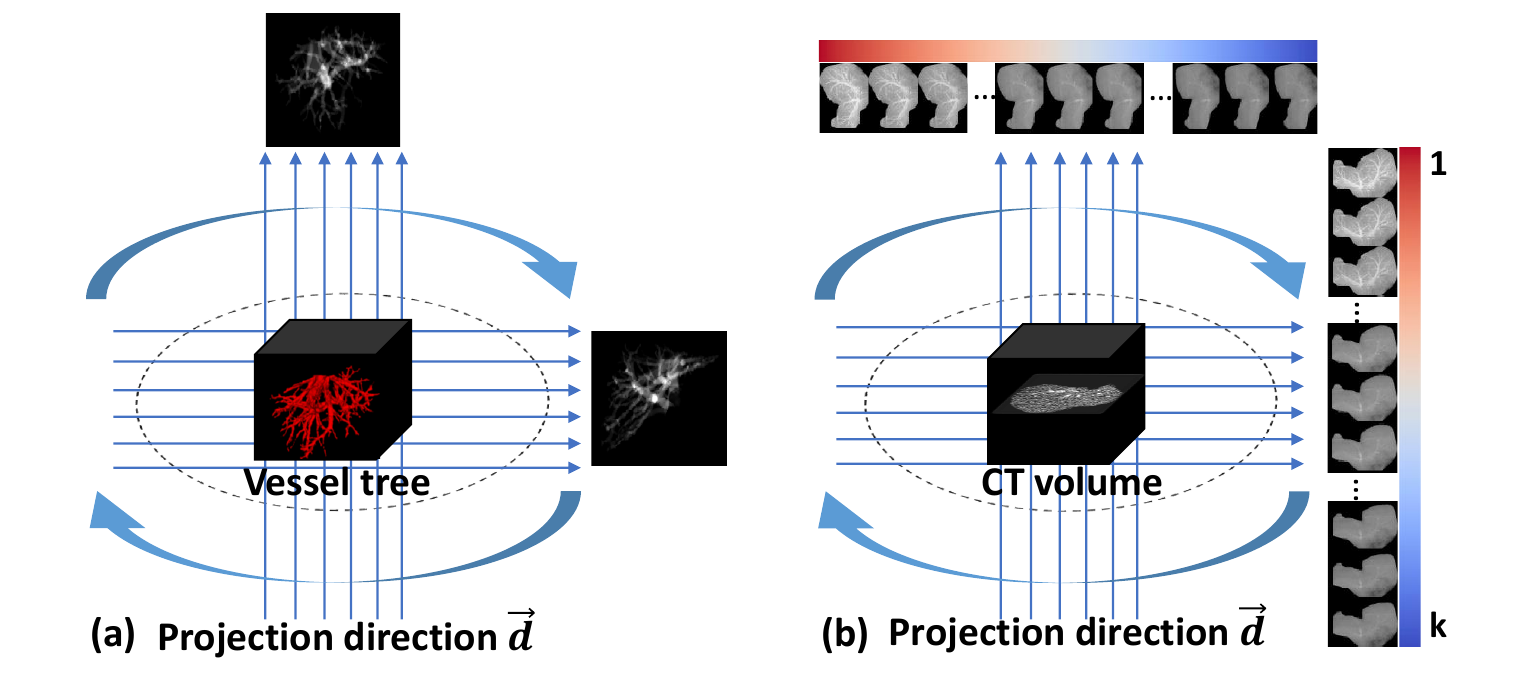}  
  \caption{\small
  \textbf{(a) Ground truth IP:} We include all vessel tree slices in the \textsl{integral projection (IP)} and accumulate them along projection directions $\overrightarrow{d}$.
  \textbf{(b) Condition top-k MIP:} We include all CT slices in the \textsl{top-k maximum intensity projection (top-k MIP)} and keep the \textsl{top-k} maxima along projection directions $\overrightarrow{d}$;  
  }
  \label{fig:msMIP_IPs}
\end{figure}
For obtaining the ground truth \textsl{integral projection (IP)}, similar to the CT forward projection \cite{b17}, we assume a bundle of parallel rays $R^{\overrightarrow{d}}$ across different directions $\overrightarrow{d}$, penetrating the ground truth binary liver-vessel tree, as in \fig{msMIP_IPs}(a). 
The \textsl{IP} is the sum of the attenuation coefficient $\mu$ ($\mu{=}1$ for binary liver-vessel trees) along a unit path $\Delta l^{\overrightarrow{d}}$, which is equivalent to the logarithm of the ratio between the ray intensity $R^{\overrightarrow{d}}$ after the attenuation by the liver vessel tree, and the initial ray intensity $R^{\overrightarrow{d}}_0$:   
\begin{align}
    {\mathbf{x}}_0 \myeq \sum_{\Delta l^{\overrightarrow{d}}} \mu = -\ln \left( \frac{R^{\overrightarrow{d}}}{R^{\overrightarrow{d}}_0} \right).
    \label{eq:IPs}
\end{align}
For computing the condition $\mathbf{c}$ to be input to the latent diffusion model we use \textsl{top-k MIP}.
Different from the integral attenuation in \Eq{IPs}, \textsl{top-k MIP} keeps the \text{top-}k maxima along a ray path $\Delta l^{\overrightarrow{d}}$, as shown in \fig{msMIP_IPs}(b):
\begin{align}
    \mathbf{c} \myeq \text{top-}k\left\{ \max\left( \Delta l^{\overrightarrow{d}}_i  \mu,  \Delta l^{\overrightarrow{d}}_{i+1} \mu, \dots, \Delta l^{\overrightarrow{d}}_{n} \mu \right) \right\},
    \label{eq:msMIP}
\end{align}
where $n$ is the number of CT voxels located along a ray, and $\Delta l^{\overrightarrow{d}}_i \mu$ is the value of the $i$-th CT voxel along the ray.

\subsection{\textsl{Top-k MIP} conditioning latent diffusion model}
We rely on a latent diffusion model \cite{b19} conditioned on the \textsl{top-k MIP}, $\mathbf{c}$, for recovering the denoised liver vessel \textsl{IPs}, $\hat{\mathbf{x}}_{0}$. 
We encode $\mathbf{c}$ into a latent vector via interpolation and convolution (yellow block in \fig{network}), into $\mathbf{c}^\prime$.
Following \cite{b19}, we also encode the \textsl{IP}, $\mathbf{x}_0$, via a pre-trained Kullback-Leibler (KL)-divergence auto-encoder (green blocks in \fig{network}) to obtain $\mathbf{x}^\prime_0$.

Standard in the diffusion forward process, we iteratively add Gaussian noise $\epsilon{\sim}\mathcal{N}(\mathbf{0}, \mathbf{I})$ to $\mathbf{x}^\prime_0$.
We aim to denoise the noisy inputs $\mathbf{x}^\prime_t, \text{ for }t{\in}\{T, .., 0\}$ by estimating the noise $\epsilon_\theta$, modelled as a U-Net \cite{b19} (depicted in blue in \fig{network}).
For this, we optimize the parameters of the model, $\theta$, by minimizing the loss over time steps, $t$:
\begin{align}
    \mathcal{L}_t(\mathbf{x}^\prime_0, \epsilon, \theta) &= \lVert \epsilon_\theta(\mathbf{x}^\prime_t \mid \mathbf{c}^\prime, t) - \epsilon \rVert _2^2. 
    \label{eq:loss}
\end{align}

Once the model is trained, given the encoded \textsl{top-k MIP} condition $\mathbf{c}^\prime$, and the input Gaussian noise $\mathbf{x}^\prime_T$, we gradually denoise the noisy input $\mathbf{x}^\prime_t, t{\in}\{T, .., 0\}$ to the estimate of the encoded \textsl{IP}, $\hat{\mathbf{x}}^\prime_0$. 
Given the estimate of the encoded \textsl{IP}, $\hat{\mathbf{x}}^\prime_0$, we decode this via the pre-trained KL-divergence auto-encoder (green block in \fig{network}) into an estimate of the \textsl{IP}, $\hat{\mathbf{x}}_0$.

\subsection{Post-processing for artifact suppression}
Given the estimated \textsl{IP}s of the vessel tree, $\hat{\mathbf{x}}_0$, we reconstruct the 3$D$ liver vessel tree via filtered back projection (FBP) \cite{b17}.
Given that, minor inconsistency in the projection domain can cause severe stripe artifacts in the reconstructed images, we employ a simple optimization to suppress these artifacts.
Specifically, assume $\mathbf{T}$ initialized by CT image is a reconstructed liver vessel tree without artifact, and given a matrix $\mathbf{A}^{\overrightarrow{d}}$ recording the voxel indices penetrated by rays under different projection directions $\overrightarrow{d}$, we impose a projection consistency.
Concretely, to dilute the irrelevant background introduced by the initialization of $\mathbf{T}$,  we enforce that the projection of the reconstructed tree $\mathbf{T}$ on a certain projection direction encoded in $\mathbf{A}_{\overrightarrow{d}}$ should be as close as possible to the estimated \textsl{IP} under that projection direction, $\hat{\mathbf{x}}^{\overrightarrow{d}}_0$: 
\begin{align}
    \mathbf{T} = \arg\min_{\mathbf{T}} \sum_{\overrightarrow{d}}  \rVert \mathbf{A}^{\overrightarrow{d}} \mathbf{T} - \hat{\mathbf{x}}^{\overrightarrow{d}}_0 \rVert_2^2.
    \label{eq:opt}
\end{align}
In practice, performing this optimization $10$ times is sufficient to suppress the stripe artifacts.
Thus, we obtain the binary liver-vessel tree $\mathbf{T} \geq \text{percentile}(\mathbf{T}, 95)$. 
Finally, we perform connected region analysis to cancel small spurious predictions surrounding the segmented vessel tree.
\section{Experimental evaluation}
\label{sec:exps}

\begin{table*}
    \centering
    \caption{\small
    \textbf{Quantitative evaluation on the \textsl{3D-ircadb-01} \cite{b23} dataset}.
    Our method predicts complete liver-vessel segmentations, while having the highest \textsl{IoU}, \textsl{Sen} and \textsl{DSC} scores. 
    The lower \textsl{clDice} scores of our method compared to \textsl{nnUNet} could be due to imprecise annotations.
    Finally, there is a trade-off between complete segmentation (higher \textsl{Sen}) or accurate segmentation (higher \textsl{Spe}). 
    }
    \resizebox{.95\linewidth}{!}{%
    \begin{tabular}{lcllllllll}
    \toprule
        & Conv type & View type & \textsl{DSC} (\%) & \textsl{clDice} (\%) & \textsl{IoU} (\%) & \textsl{Sen} (\%) & \textsl{Spe} (\%) & \textsl{FLOPs} (G) & \textsl{Params} (M)  \\
    \midrule
    nnUNet\cite{b24}&3D & cross section& $58.76\pm9.89$ &$\textbf{71.46}\pm{5.67}$ & $42.31\pm10.19$ & $43.32\pm11.18$& $\textbf{100}\pm{0}$ &$2.90\times10^3$ &$30.7$  \\
    Swin UNETR\cite{b25}&3D & cross section& $57.80\pm9.93$ &$64.16\pm7.10$ & $41.31\pm9.51$ & $46.71\pm13.21$& $99.96\pm0.02$ &$6.15\times10^2$ &$62.2$  \\
\midrule
    EnsemDiff\cite{b26}&2D & cross section& $54.82\pm9.64$ &$60.61\pm9.55$ & $38.37\pm9.21$ & $40.05\pm10.45$& $99.98\pm0.02$ &$9.96\times10^2$ &$113.7$  \\
    MedSegDiff\cite{b27}&2D & cross section& $59.59\pm7.73$& $66.03\pm8.05$ & $42.85\pm7.53$ & $47.38\pm10.44$& $99.95\pm0.05$ &$1.05\times10^3$ &$136.8$  \\
    Ours &2D  & projection& $\textbf{64.25}\pm{7.19}$& $65.87\pm8.16$ & $\textbf{47.75}\pm8.00$ & $\textbf{54.03}\pm{11.23}$& $99.96\pm0.01$ &$6.83\times10^2$ &$84.5$\\
    \bottomrule
    \end{tabular}}
    \label{tab:results1}
\end{table*}

\subsection{Experimental setting}
We evaluate our proposed model on the \textsl{3D-ircadb-01} \cite{b23} dataset, consisting of $20$ CT scans. 
We exclude the vena cava from the region of interest, using liver masks to focus only the vessels within the liver.
We resize the liver-masked ROIs to $[256{\times}256{\times}256]$ for each CT scan, and project $180$ viewing directions from $0$ to $180$ degrees for each CT scan, totaling $3600$ projections.
$k$ is $32$ when creating the \textsl{top-k MIP}.

Small liver vessel annotations are incomplete in the \textsl{3D-ircadb-01} dataset \cite{b7}. 
To mitigate this, we asked clinical experts to score the completeness of the annotations based on a 5-point criterion and picked cases $(4,6,8,11,16)$ whose score $\geq4$ as the test set.
We apply leave-one-out cross-validation, such that we have 19 cases (\ie 3420 projections) for training and a single case (\ie 180 projections) for testing, per fold.

We trained \textsl{nnUNet}\cite{b24} and \textsl{SwinUNetr}\cite{b25} using 3$D$ CT sub-volumes as input, and using their default pre- and post-processing. 
For \textsl{MedSegDiff}\cite{b27} and \textsl{EnsemDiff}\cite{b26} we used 2$D$ CT slices, and ensemble segmentation results $5$ times.  
Our method does not require an ensemble.
For all the experiments, we used an NVIDIA A100 (40GB) GPU.

\subsection{Experimental results}
\begin{figure*}
  \centering
  \includegraphics[width=\linewidth]{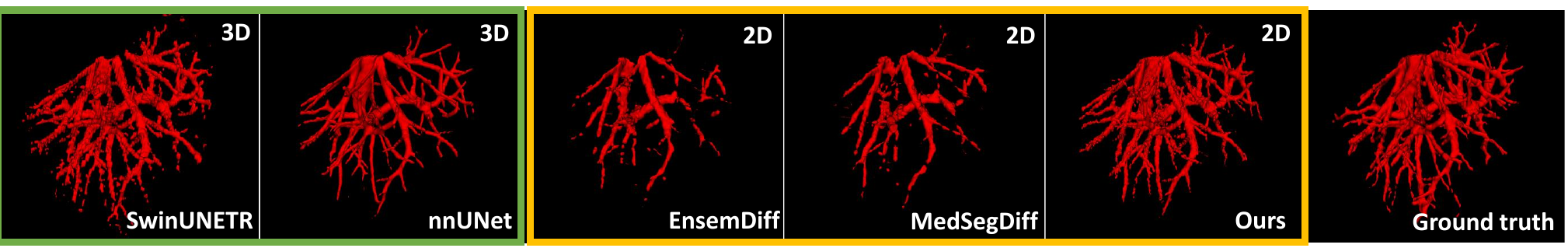}
  \caption{\small
  \textbf{Qualitative comparison on the \textsl{3D-ircadb-01}\cite{b23} dataset.}
    We highlight in green the predictions of the 3$D$ methods and in yellow the predictions of the 2$D$ methods.
    Our model makes more complete and continuous predictions, resembling the ground truth.
  }
  \label{fig:compare_IRCADB}
\end{figure*}

We report voxel-wise Dice coefficient (\textsl{DSC}), Intersection-Over-Union (\textsl{IoU}), Sensitivity (\textsl{Sen}), Specificity (\textsl{Spe}) and centerline Dice coefficient (\textsl{clDice}) \cite{b28}.
Table \ref{tab:results1} shows the quantitative evaluation, while \fig{compare_IRCADB} shows a few prediction examples.
Our method achieves the highest \textsl{DSC}, \textsl{IoU}, and \textsl{Sen} scores compared to the other baselines on the \textsl{3D-ircadb-01} dataset, demonstrating the added value of the proposed top-k MIP projections. 
\textsl{nnUNet} has a higher \textsl{clDice} score than our method. However, the vessel centerline was estimated from ground truth vessel annotations, following \cite{b28}.
Therefore, incomplete and discontinuous voxel-wise ground truth could bias the centerline extraction in \textsl{clDice} \cite{b28}. 
\fig{compare_IRCADB} shows that both our method and \textsl{nnUNet} predict well-connected vessel structures compared to the ground truth, indicating that \textsl{clDice} might not be a sensible choice when comparing segments with inconsistent connectivity.
Additionally, all methods in Table \ref{tab:results1} have \textsl{Spe} scores close to $1$, indicating that there are few false positives in liver vessel segmentation.

\vspace{-5px}
\section{Discussion and limitations}
\vspace{-5px}


\noindent\textbf{Effect of artifact suppression.}
Our model generates liver vessel \textsl{IPs} based on the intensity of the \textsl{top-k MIP}. 
However, due to memory constraints, our current design limits the model from learning correlations between different projection directions.
The optimization in \Eq{opt} compensates for this shortcoming.
This optimization suppresses the reconstruction artifact caused by projection inconsistency.
\begin{table}[t]
    \centering
    \caption{\small
    \textbf{Liver vessel \textsl{IPs} estimation with (w/) and without (w/o) artifact suppression (opt)}.
    Artifact suppression makes the vessel \textsl{IPs} more similar to the ground truth vessel \textsl{IPs}.
    }
    \resizebox{.6\linewidth}{!}{%
    \begin{tabular}{lll}
    \toprule
        & PSNR($\uparrow$) & SSIM($\uparrow$)  \\
    \midrule
    w/o opt & $13.51\pm0.78$ & $0.45\pm0.07$ \\
    w/ opt & $\textbf{15.15}\pm1.32$ & $\textbf{0.66}\pm0.05$ \\  
    \bottomrule
    \end{tabular}}
    \label{tab:IPs_quality}
    \vspace{-5px}
\end{table}

\begin{figure}[t]
  \centering
  \includegraphics[width=\linewidth]{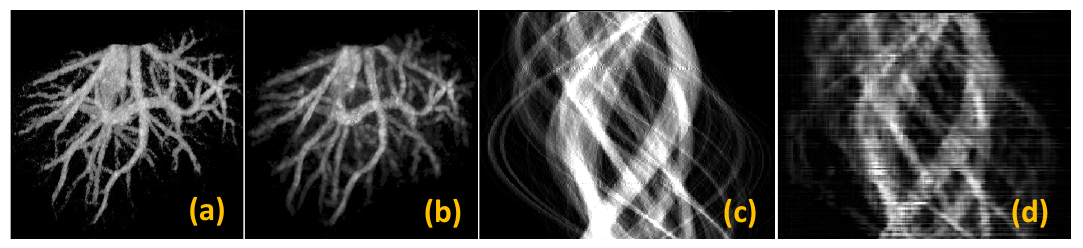}
  \caption{\small
  \textbf{Vessel reconstruction with\slash without artifacts suppression.}
  (a) reconstructed vessel tree with artifact suppression; 
  (b) reconstructed vessel tree without artifact suppression; 
  (c) full-view projection of a slice of the reconstructed vessel trees with artifacts suppression;
  (d) full-view projection without artifacts suppression. 
  Artifact suppression improves the consistency between different projection views. 
  The vertical axis in (c) and (d) is the projection view.
  }
  \label{fig:artifact}
\end{figure}
Table ~\ref{tab:IPs_quality} tests the added value of this artifact suppression optimization. 
The \textsl{IPs} of the reconstructed vessel tree with artifact suppression have higher \textsl{PSNR} \cite{b29} and \textsl{SSIM} \cite{b29}, indicating that the optimized \textsl{IPs} have a similar appearance to the ground truth.
Full-view projections in \fig{artifact}(c)-(d) also show that the vessel \textsl{IPs} consistency between projections is improved by the artifact suppression. 

\smallskip\noindent\textbf{Model limitations.}
\begin{figure}
  \centering
  \includegraphics[width=.7\linewidth]{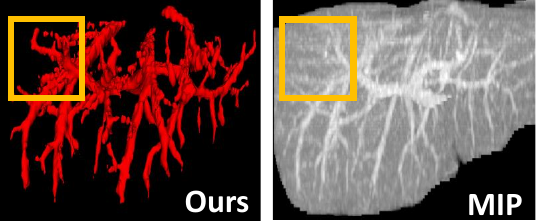}
  \caption{\small
  \textbf{Failure analysis.}
  Correctly segmenting low-contrast CT images is challenging for our method.
  }
  \label{fig:limit}
\end{figure}
Our proposed model can handle most cases, where the contrast is within normal ranges, in the liver vessel area.
However, segmenting extremely low-contrast vascular regions, as shown in the yellow box in \fig{limit} remains challenging for our model.

\smallskip\noindent\textbf{Possible improvements.}
Strengthening the correlation between projection views and learning a dynamic projection for the \textsl{top-k} MIP are the two improvement directions. 
Currently, each view is processed independently, and thus we cannot test using the standard MIP. In the standard MIP the depth information is collapsed to a single maximum value, and cannot be recovered.
A model that considers the correlation between projection views, will make it possible to use of the standard MIP, instead of the \textsl{top-k} MIP.

\vspace{-5px}
\section{Conclusion}
\vspace{-5px}
We propose a principled way of incorporating 3$D$ liver-vessel topology in 2$D$ diffusion models for liver-vessel segmentation.
Accordingly, we draw inspiration from the physics involved in the 3$D$ CT reconstruction.
Concretely, we propose to condition a latent diffusion model on \textsl{top-k maximum intensity projections} of the CT cross-sectional view.
Our proposed model achieves competitive results compared to the existing baselines, validating our approach.

\vspace{-5px}
\section{Compliance with Ethical Standards}
\vspace{-5px}
This research study was conducted retrospectively using human subject data made available in open access by IRCAD, Strasbourg, France. Ethical approval was not required as confirmed by the license attached with the open access data.

\section{Acknowledgements}
This work was supported by China Scholarship Council under Grant 202108310010.

\bibliographystyle{IEEEbib}
\bibliography{refs}

\end{document}